\documentclass[twoside]{article}

\usepackage[accepted]{art}
\usepackage[T1]{fontenc}
\usepackage{hyperref}
\usepackage{url}
\usepackage{booktabs}
\usepackage{amsmath, amssymb}
\usepackage{color}
\usepackage{graphicx}
\usepackage{mathrsfs}
\usepackage{enumitem}
\usepackage{array}
\usepackage{colortbl}
\usepackage{ctable}
\usepackage{multirow}
\usepackage{xcolor}
\usepackage[flushleft]{threeparttable}
\newcommand\sm[1]{\textcolor{black}{#1}}

\begin{document}
	
\twocolumn[
	
	\arttitle{Controlling Information Capacity of Binary Neural Network}
	
	\artauthor{ Dmitry Ignatov \And Andrey Ignatov }
	
	\artaddress{ Russian Research Center\\
		Huawei Technologies, Russia\\
		\texttt{dmitri.ignatov@gmail.com} \And  Department of Computer Science\\
		ETH Zurich, Switzerland \\
		\texttt{andrey.ignatoff@gmail.com} } ]
	
	\begin{abstract}
	Despite the growing popularity of deep learning technologies, high memory requirements and power consumption are essentially limiting their application in mobile and IoT areas. While binary convolutional networks can alleviate these problems, the limited bitwidth of weights is often leading to significant degradation of prediction accuracy. In this paper, we present a method for training binary networks that maintains a stable predefined level of their information capacity throughout the training process by applying Shannon entropy based penalty to convolutional filters. The results of experiments conducted on SVHN, CIFAR and ImageNet datasets demonstrate that the proposed approach can statistically significantly improve the accuracy of binary networks.
	\end{abstract}
	
	\section{Introduction}
	\label{Introduction}
	With a rapid evolution of mobile devices from simple tools for voice and data transfer to multifunctional intelligent gadgets performing multiple deep learning tasks like image classification, object detection or natural language processing, there is a substantial need in fast, power-efficient and robust neural networks. One promising approach to this problem is reducing the bitwidth of weights, which in extreme cases leads to emission of binary networks, where commonly used floating-point weights are replaced with binary ones. Such reduction results in up to 32 times smaller network size and more efficient inference on CPUs and specialized hardware~\cite{Ignatov_A18, Ignatov_A19}. However, binarization also leads to a substantial accuracy drop caused by the reduction of information capacity in the convolution filters. The latter one can be measured by Shannon information entropy ($H$) according to the basic principles of the information theory.
	
	In this paper, we propose a novel solution for the problem of restricted performance of binary networks that is controlling and stabilizing  of their information capacity. The proposed approach ensures the best representation of information in the convolutional filters, which leads to a higher overall prediction accuracy.
	
	Our main contributions are as follows:
	\begin{itemize}[label=$\bullet$]
		\item For the first time, we propose the concept of information capacity regularization in convolutional neural networks.
		\item A new regularization technique is developed: Shannon entropy based information loss penalty for binary convolutional neural networks.
		\item We show that maintaining the upper-level value of information entropy in convolutional filters during the training process leads to a higher prediction accuracy of binary neural networks.
	\end{itemize}
	
	The rest of the paper is organized as follows. Section~\ref{Background} gives an overview of the related works. Section~\ref{Maintaining} describes the proposed approach in detail, and Section~\ref{Empirical} provides the experimental results obtained on several common machine learning problems. Section~\ref{Conclusion} concludes this work.
	
	\section{Background}\label{Background}
	In this section, we review the existing techniques related to binary neural networks and application of the entropy-based approaches in the field of machine learning.
	
	\subsection{Binary Neural Networks} Training and inference with deep neural networks usually consume a large amount of computational power, which makes them hard to deploy on mobile devices. Recent network binarization approaches~\cite{Courbariaux_2015, Courbariaux, Hubara16, Hubara, Rastegari, Zhou} offer an efficient solution for these tasks by using binary weights and fast bitwise operations that are radically accelerating the computations and reducing the size of the network by up to 32 times, however, at the cost of lower prediction accuracy. According to the theoretical analysis~\cite{Rastegari}, the binarization of weights and quantization of activation functions can bring 2 to 58 times speedup. The convolutions with binarized weights require only addition and subtraction operations (without multiplications) that results in 2 times reduction of computation time, while additional binarization of activations, and usage of XNOR and POPCOUNT operations can provide up to 58 times speedup~\cite{Courbariaux}.
	
	In works~\cite{Courbariaux_2015, Courbariaux, Hubara}, the authors proposed constraining weights and activations to binary values $\in \{1, -1\}$ and using efficient XNOR and POPCOUNT operations instead of matrix multiplications. Built on a similar idea, XNOR-Net~\cite{Rastegari} introduces a channel-wise scaling factor to improve approximation of full-precision weights and stores weights between layers in real-valued variables. Replacing the channel-wise scaling factor with one scalar for all filters of each layer and focusing on different bitwidth of weights and activations is the main idea of DoReFa-Net~\cite{Zhou} that shows the effectiveness of the binary network with 4-bit activations. The quantization of activations in binarized U-Net~\cite{Tang} with varied bitwidth reduces the memory consumption by up to 4 times without compromising the performance. An ambitious attempt to use only binary weights is taken in ABC-Net~\cite{Lin3} that is essentially reducing the drop in top-1 accuracy on the ImageNet dataset. By providing 3 to 5 binary weight bases to approximate full-precision weights, ABC-Net demonstrates an increased information capacity, as well as an enlarged size and complexity.
	
	While the previously proposed approaches are achieving a significantly reduced computational load and smaller network size due to the restricted weight bitwidth, this also leads to essential drop of prediction accuracy. Finding a way to accurately train binary neural networks still remains an unsolved task. One natural way to solve this problem is to measure and control the quantity of information in the binary network that can be done, for example, with a classical measure of information quantity~--- Shannon information entropy.
	
	\subsection{Entropy-Based Learning Methods} Shannon information entropy ($H$) is defined as an expected amount of information in a random variable~\cite{Shannon}, and is widely used, together with its successors, as a measure of information value in different fields, including chemistry, medicine~\cite{Ignatov2005,Ignatov2012,Ignatov2001}, robotics and machine learning. In deep belief networks, the maximum entropy learning algorithm provides better generalization capability than the maximum likelihood learning approach, ensuring a less biased distribution and robust to over-fitting predictive model~\cite{Jing}. Shannon entropy inspired the creation of cross-entropy loss function~\cite{Richard} that is commonly used for training various networks performing data classification. Another successor is the Kullback-Leibler divergence~\cite{Kullback} that is employed for training deep belief networks~\cite{Lin}, where maximization of the model parameters entropy ensures a more efficient generalization. Maximum entropy principle is also used for maximization of the expected reward in reinforcement learning~\cite{Norouzi,Shen,Xiong} to define the probability of a trajectory and to choose a near-optimal expert policy~\cite{Audiffren, Ziebart}.
	
	According to the principle of maximum entropy $H$, the better generalization is achieved when the input data distribution is as uniform and unbiased as possible when dealing with estimation problems given incomplete knowledge~\cite{Japkowicz}. Thus, increasing the entropy of the input data by adding random noise improves the robustness to over-fitting on small data sets~\cite{Amos, Japkowicz}. Multi-sensor data fusion with entropy maximization also ensures an effective imaginary reconstruction~\cite{Shkvarko}. Maximum entropy principle gives the ability to extract the learned features from the input data and provides a better prediction accuracy for deep neural networks~\cite{Finnegan}. High information quantity in the form of the highest homogeneity of samples in child nodes is maintained by entropy-based information gain ratio and Gini impurity measures in such nonparametric supervised learning methods as decision tree~\cite{Quinlan} and decision stream~\cite{Ignatov}.
	
	Entropy maximization methods are used to train generative networks for texture synthesis~\cite{Loaiza-Ganem} and for reconstruction of macroeconomic models~\cite{Hazan}. It was shown that penalizing the entropy of the network's output distribution improves the exploration in reinforcement learning, acts as a strong regularizer in supervised learning~\cite{Pereyra}, and improves medical image segmentation~\cite{Hu}. In this paper, we make an attempt to go further and propose a new regularization approach for controlling information entropy of the weight distribution in the convolutional filters of binary networks.
		
	\section{Information Loss Penalty}
	\label{Maintaining}
	In this section, we introduce an approach for controlling information capacity of binary neural networks with Shannon information based loss penalty. First, we describe the calculation of the entropy $H$ for binary convolutional filters, and then present the information loss penalty for networks with binary weights backed by real-valued variables.
	
	The sum $S(\cdot)$ of absolute values of weights ${W_f}$ in the convolutional filter ${f}$ and the difference $\Delta(\cdot)$ between the quantity of positive ($+1$) and negative ($-1$) values of ${W_f}$ are defined for every convolutional filter as:
	\begin{equation}
	S({W_f}) = \sum^n_{i = 1}{|{w_{f_i}}|}\;,
	\end{equation}

	\begin{equation}
	\Delta({W_f}) = \sum^n_{i = 1}{{w_{f_i}}},
	\end{equation}

	where $n$ denotes the size of tensor ${W_f}$.
	These values are used to calculate the probability density functions for positive and negative weights ${W_f}$ in the filter  according to:

	\begin{equation}
		P({W_f})=\frac{S({W_f}) +\Delta({W_f})}{2\cdot S({W_f})}\,,
	\end{equation}
	\begin{equation}
		N({W_f})=\frac{S({W_f}) -\Delta({W_f})}{2\cdot S({W_f})}\,.
	\end{equation}

	The Shannon information entropy of the binary weight distribution in convolution filter is then computed using the above density functions $P(\cdot)$ and $N(\cdot)$ as:
	
	\begin{equation}
	\label{equ:h:ft}
	\begin{split}	
	H_{f}({W_f}) = -(P({W_f}) \cdot Log_2 P({W_f}) + \\
	N({W_f}) \cdot Log_2 N({W_f}))\;.
	\end{split}	
	\end{equation}	
		
	The mean information entropy for all convolutional filters in the network with binary weights ${W^B}$ can be obtained with:
	
	\begin{equation}
	\label{equ:h:av}
	\overline{H}_{f}({W^B}) = \frac{\sum^N_{{f = 1}}{H_f({W^B_f})}}{N}\;,
	\end{equation}
	
	where $N$ denotes the total number of filters, and ${W^B_f}$ is a tensor with binary weights corresponding to filter ${f}$. The estimation of the time complexity of function $\overline{H}_{f}(\cdot)$ for conventional deep neural network architectures is $O(N)$.
	
	Following~\cite{Courbariaux}, to binarize a floating-point network we constrain the weights to $+1$ and $-1$, which is advantageous from a hardware perspective, and in order to transform the real-valued variables into discrete values, we use the deterministic binarization function ${B}$:
	
	\begin{equation}
	\label{equ:h:bin:weights}
	\begin{split}
	W^B \leftarrow {B}({W^R}) \triangleq Sign({W^R}) \triangleq
	{W^R} \mapsto  \\ W^B :
	{W^B_i} = \left \{
	\begin{tabular}{cc}
	$+1$ & if ${W^R_i} \geq 0$,
	\\ $-1$ & otherwise,
	\end{tabular}
	\right.
	\end{split}
	\end{equation}
	
	where the calculated binary weights ${W^B}$ are used in both forward and backward passes of the network computations, and the original real-valued weights ${W^R}$ are utilized in the update phase of the training procedure.
	
	The previously introduced information entropy (Eq.~\ref{equ:h:av}) was defined only for binary weights, while the network optimization is performed on floating-point variables. Though binary weights can be substituted with ${W^B} = Sign({W^R})$, in this case we will get a non-differentiable loss function. To avoid this problem and provide differentiable $\overline{H}_{f}(.)$, we propose approximating the $Sign$ function of ${W^R}$ with:
	
	\begin{equation}
	\begin{split}	
	{\widehat{W}^B} \leftarrow Tanh(10^k \cdot {W^R}) \triangleq {W^R} \mapsto \\ {\widehat{W}^B} :
	{\widehat{W}^B_i} = Tanh(10^k \cdot {W^R_i})\,,
	\end{split}	
	\end{equation}
	
	where
	$||{W^B} - {\widehat{W}^B}|| < 10^{-8}$ for $k=5$ based on the experimental results. Using this approximation of binary weights ${\widehat{W}^B}$, we can estimate the information loss as an absolute difference between the expected information entropy $H_{e}$ and the actual one $\overline{H}_{f}({\widehat{W}^B})$:

	\begin{equation}
	\label{equ:h:loss}	
	H_{Loss}({\widehat{W}^B}) = ||H_{e} - \overline{H}_{f}({\widehat{W}^B})||\;,
	\end{equation}
	
	where $H_e$ provides the control over the network information capacity, and its biggest value for the binary distribution ($H_e = 1$) corresponds to the maximum information capacity.
	The resulting information loss penalty is added to the overall loss function in a conventional way:	

	\begin{equation}
	\label{equ:h:lss}
	Loss\;\mathrel{+}= \lambda\cdot H_{Loss}({\widehat{W}^B})\;,
	\end{equation}
	
	and is used in the back-propagation algorithm for network training.
	
	\section{Empirical Verification}
	\label{Empirical}
		To evaluate the performance of the information loss penalty and to explore its properties, we conducted several experiments with binary networks on common machine learning problems. Below we are examining the impact of $H_{e}$ on the network's performance and information capacity, and compare the obtained accuracy values (mean and 95\% confidence interval) with the results of the standard binary and full-precision networks.

	\subsection{Datasets}
	\label{Dataset}
	
	\begin{itemize}[label=$\bullet$]
		\item \textit{CIFAR-10 and CIFAR-100  image classification datasets~\cite{Krizhevsky}}: \small{10 and 100 classes, respectively, 32$\times$32 px images, 50K training and 10K validation samples.}\normalsize\footnote{https://www.cs.toronto.edu/\textasciitilde kriz/cifar.html}
		\smallskip
		\item \textit{SVHN a real-world digit image classification dataset~\cite{Netzer}}: \small{10 classes, 32$\times$32 px images, 73K training and 26K validation samples.}\normalsize\footnote{http://ufldl.stanford.edu/housenumbers}
		\smallskip
		\item \textit{ImageNet ILSVRC12 classification dataset~\cite{Russakovsky}}: \small{1000 classes, 224$\times$224 px images, 1.28M training and 50K validation samples.}\normalsize\footnote{http://www.image-net.org}
	\end{itemize}
	
	\subsection{Training Process}		
	All experiments were performed on Nvidia GK110B GPUs for both full-precision and binary networks. The models were trained to minimize cross entropy loss function additionally augmented with the information loss penalty (Eq.~\ref{equ:h:loss}) in the corresponding experiments.
	
	Following~\cite{Huang}, all networks were trained using stochastic gradient descent. On SVHN and  CIFAR datasets, the networks were trained for 40 and 300 epochs, respectively, using batch sizes of 64 and 128 for full-precision and binary networks. The initial learning rate was set to 0.1, and was divided by 10 at 50\% and 75\% of the total number of training epochs. On ImageNet dataset, the images were resized to 224$\times$224 px, and the models were trained for 90 epochs with a batch size of 256. The learning rate was initially set to 0.1, and was decreased by 10 times at epochs 30 and 60. A weight decay of $10^{-4}$~\cite{Gross} and a Nesterov momentum of 0.9~\cite{Sutskever} were used without damping. 
	We utilized DoReFa-Net binarization technique for weights quantization~\cite{Zhou}, adapting its implementation\footnote{https://github.com/zzzxxxttt/Pytorch\_DoReFaNet} for our tasks, and keeping full-precision first and last layers~\cite{Bulat, Hubara16, Zhou}. A 4-bit uniform quantization is applied to activations~\cite{Lin2} starting from the outputs of the first layer.
	
	To ensure a fair comparison between full-precision, binary networks and binary networks with the information loss penalty, we are using identical experimental settings in all three cases, including data preprocessing and optimization settings. For the experiments, we used five publicly available network architectures: for ImageNet classification --- PyTorch\footnote{https://pytorch.org (PyTorch v. 1.0, Cuda v. 10.0)} implementation of DenseNet-121~\cite{Huang}, ResNet-18~\cite{He0}, Inception~v3~\cite{Szegedy} and AlexNet~\cite{Krizhevsky12} from TorchVision package, for other datasets --- pre-activation ResNet-18\footnote{https://github.com/kuangliu/pytorch-cifar}~\cite{He}. An auxiliary classifier in the binary Inception network was kept full-precision. Binarization leads to 32$\times$ compression of the hidden layers and a commensurate reduction of the model memory footprint (Table~\ref{table:net:size}).
	
	\begin{table}[h!]
		\begin{threeparttable}
			\caption{Memory consumption of the inference models.}
			\label{table:net:size}
			\def\arraystretch{1.2}
			\newcommand{\GC}{\cellcolor{gray!10}}
			\newcolumntype{A}{ >{\centering\arraybackslash} m{0.9cm}}
			\newcolumntype{B}{ >{\centering\arraybackslash} m{1.1cm}}
			\newcolumntype{C}{ >{\centering\arraybackslash} m{1.9cm}}
			\newcolumntype{D}{ >{\centering\arraybackslash} m{1.0cm}}
			\begin{tabular}{m{2.0cm}| A B C D}\arrayrulecolor{gray}
				\arrayrulecolor{gray}
				\hline		
				\multicolumn{1}{c|}{\multirow{2}{*}{Model}} &
				\multirow{2}{*}{\parbox{29pt}{\centering{Output \\ size}}} &
				\multicolumn{1}{|c|}{\multirow{2}{28.0pt}{\mbox{\# para-} \mbox{meters}}} &
				\multicolumn{2}{c}{Memory footprint, MB}
				\\\cline{4-5} & & \multicolumn{1}{|c|}{} & \mbox{Full-precision} & \multicolumn{1}{|c}{Binary$^*$} 		 		
				\\\specialrule{.1em}{0em}{0em} \multirow{2}{35pt}{\mbox{Pre-activation} \mbox{ResNet-18}} & 10 & 11.2M  & 44.7 & 1.4
				\\ &\GC 100 &\GC 11.2M &\GC 44.9 &\GC 1.6
				\\\cline{1-1}\mbox{DenseNet-121} & 1000 & 8.0M & 32.4 & 1.3	
				\\\cline{1-1}\mbox{ResNet-18} &\GC 1000 &\GC 11.7M &\GC 46.8 &\GC 3.6
				\\\cline{1-1}\mbox{Inception-v3} & 1000 & 27.2M & 108.9 & 11.6
				\\\cline{1-1} \mbox{AlexNet} &\GC 1000 &\GC 61.1M &\GC 244.4 &\GC 23.5 \\ \hline
			\end{tabular}
			\begin{tablenotes}
				\item $^*$ The first and the last layer are full-precision.
			\end{tablenotes}
		\end{threeparttable}	
	\end{table}
	
	 During the training process, using additional information loss penalty resulted in computational overhead of 6.3 $\pm$ 0.8\%.
				
	\subsection{Classification Results}
	\label{Results}
\begin{table*}
	\centering		
	\caption{The top-1 accuracy and 95 \% confidence intervals for full-precision and binary neural networks on four machine learning datasets.}
	\label{table:bin:accuracy}
	\def\arraystretch{1.2}
	\newcolumntype{B}{ >{\centering\arraybackslash} m{1.7cm} }
	\newcolumntype{S}{ >{\centering\arraybackslash} m{1.45cm} }
	\begin{tabular}{m{1.9cm} m{0.25cm} B B B B S S S}
		\arrayrulecolor{gray}
		\hline		
		\multicolumn{1}{c|}{\multirow{2}{*}{Model}}
		& \multicolumn{1}{c|}{\multirow{2}{*}{$H_{e}$}}
		& SVHN
		& CIFAR-10
		& \multicolumn{1}{c|}{CIFAR-100}
		& \multicolumn{4}{c}{ImageNet 2012}
		\\ \multicolumn{1}{c|}{} &  \multicolumn{1}{c|}{}& \multicolumn{3}{c|}{Pre-activation ResNet-18} & \mbox{DenseNet-121} & \mbox{ ResNet-18} & \mbox{Inception-v3} & \mbox{ AlexNet}
		\\ \specialrule{.1em}{0em}{0em}	
		\multirow{1}{1em}{\mbox{Full-precision}}& \multicolumn{1}{c}{ --- }& \mbox{96.45 $\pm$ 0.11} & \mbox{95.23 $\pm$ 0.20} & \mbox{76.78 $\pm$ 0.19} & \mbox{76.4~\cite{Huang}} & \mbox{69.3~\cite{Rastegari}} & \mbox{78.8~\cite{Szegedy}} & \mbox{56.6~\cite{Rastegari}} \\		
		\hline	
		\multirow{3}{4em}{Binary} & \multicolumn{1}{c}{ --- } & \mbox{96.10 $\pm$ 0.08}& \mbox{93.43 $\pm$ 0.19} & \mbox{73.07 $\pm$ 0.18} & 65.12 & 59.26 & 72.61 & 53.36
		\\ & \multicolumn{1}{c}{0.97}  & \mbox{\textbf{96.27$\pm$0.07}} & \mbox{\textbf{93.82$\pm$0.17}} & \mbox{\textbf{73.48$\pm$0.16}} & \textbf{67.14} & \textbf{61.36} & \textbf{73.83} & \textbf{54.07}
		\\ & \multicolumn{1}{c}{1.00}  & \mbox{96.14 $\pm$ 0.08} & \mbox{93.51 $\pm$ 0.19} & \mbox{73.20 $\pm$ 0.17} & --- & --- & --- & ---
		\\	\hline
	\end{tabular}
\end{table*}

	\begin{figure*}
		\centering{
			\includegraphics[width=15.5cm]{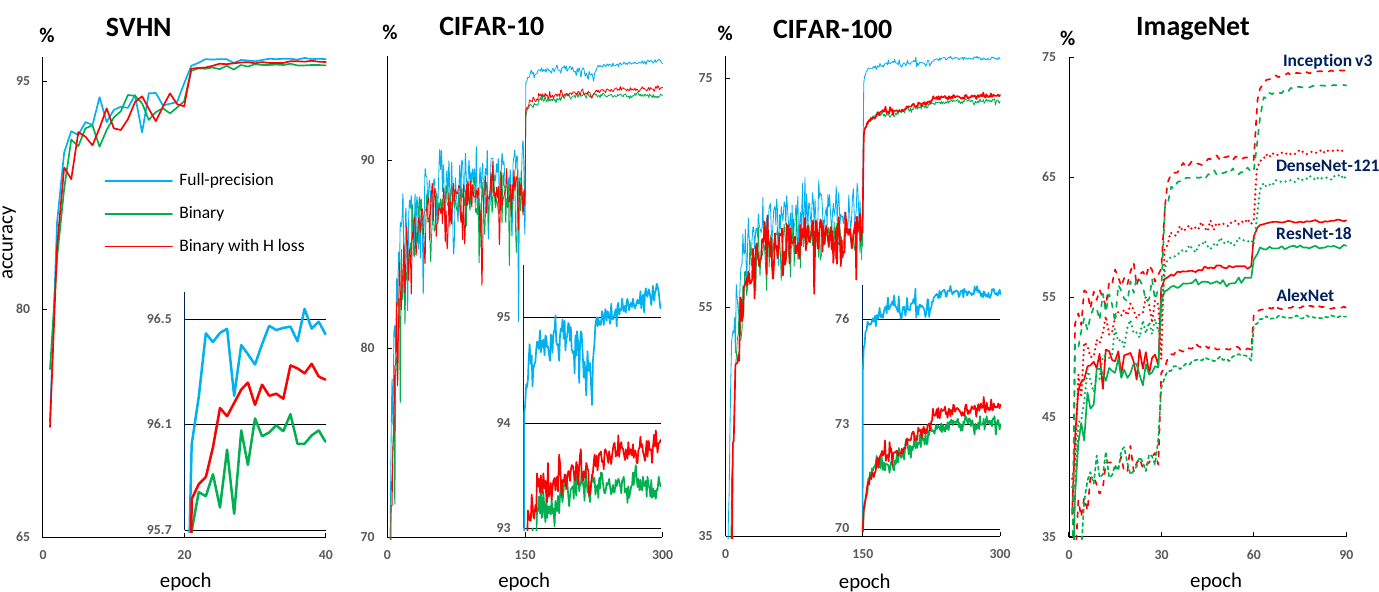}
			\caption{Training curves for full-precision and binary networks with and without the information loss penalty on four machine learning datasets.}\label{fig:curves}}
	\end{figure*}
	\begin{figure*}
		\centering{
			\includegraphics[width=12.5cm]{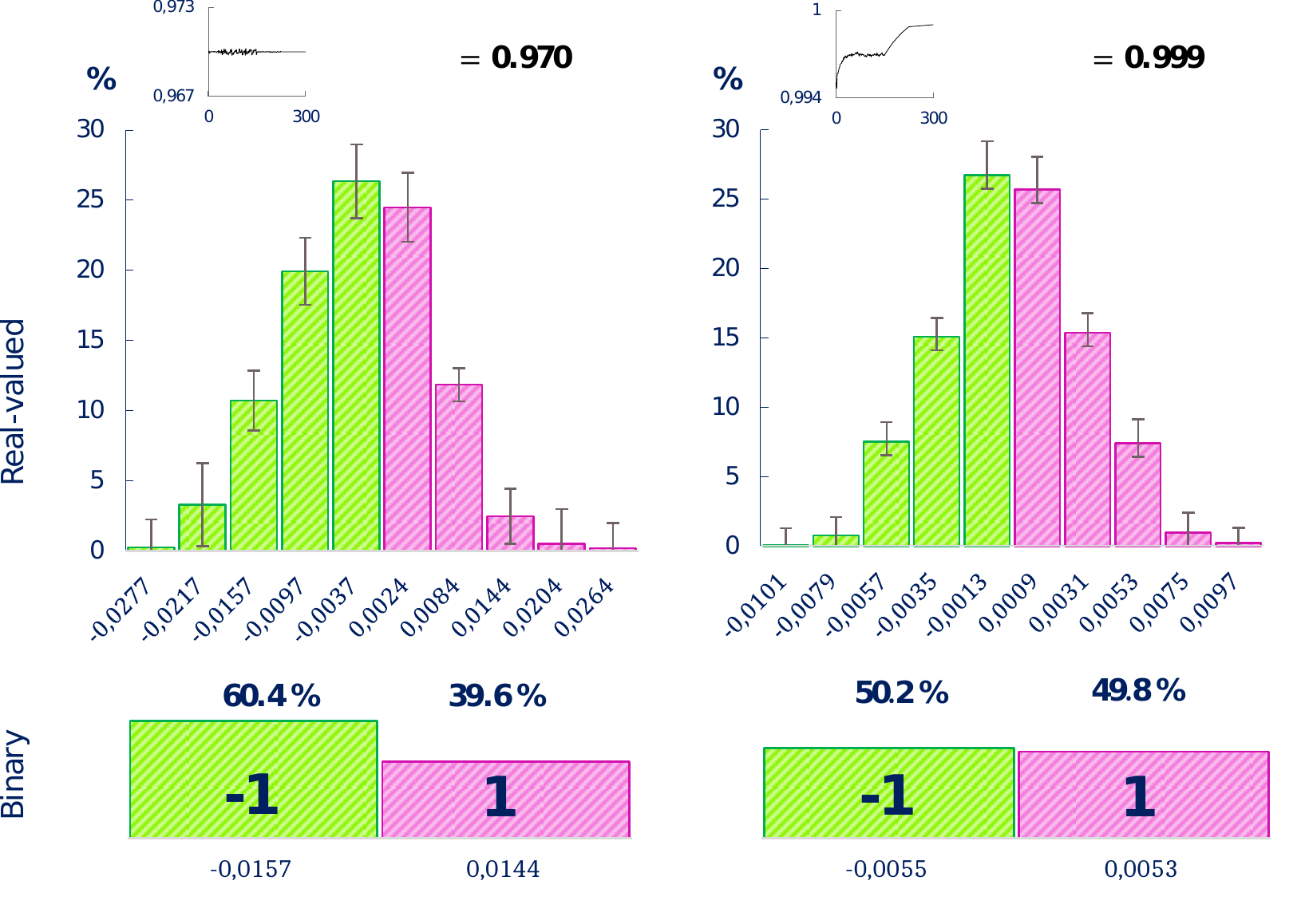}
			\caption{Histograms of real-valued (top) and binary (bottom) representations of weights ($\pm$\,95\,\% confidence interval) in binary pre-activation ResNet-18 model trained with $H_e=0.97$ (left) and $H_e=1.00$ (right) on CIFAR-100 dataset.}
			\label{fig:weighs}}
	\end{figure*}

	First, we conducted a series of preliminary experiments aimed at
	determining the optimal values of parameters $\lambda$ and $H_{e}$,
	where the accuracy of pre-activation ResNet-18 binary network was
	assessed on the validation subsets of SVHN, CIFAR-10 and CIFAR-100 datasets. In the first set of experiments, the value of the expected information entropy was kept constant and equal to its maximum value ($H_{e} = 1$), while the value	of $\lambda = 10^{-n}$ was varied between $10^{-7}$ and $1$ with a step size of n = 1, resulting in the optimal value $\lambda = 10^{-4}$.
	
	In the second set of experiments, we used the obtained value of $\lambda$, but varied the target entropy $H_e$ between 0.8 and 1 with a step size of 0.01. The results demonstrated that the optimal value of $H_e$ is equal to $0.97$, and in the next experiments we used the obtained values of $H_e$ and $\lambda$ to get the final results. We also trained pre-activation ResNet-18 network with the highest target entropy ($H_e=1$) to analyze its performance in this case.
	
	Table~\ref{table:bin:accuracy} summarizes the classification results	obtained on SVHN, CIFAR-10, CIFAR-100, and ImageNet 2012 datasets with full-precision networks and binary networks trained with common parameters and with additional information loss penalty. On SVHN and CIFAR datasets, the networks with binary weights show an accuracy drop of 0.35-3~\% compared to their full-precision versions that corresponds to the results of~\cite{Fromm, Hubara, Zhou}.
	Augmenting the target loss with the information loss penalty and aiming at the highest possible entropy value ($H_e = 1$) is leading only to a slight improvement of the accuracy by 0.04-0.13\%	in case of pre-activation ResNet model. Lowering the target entropy $H_e$ to 0.97 resulted in larger accuracy improvements: 0.17\%, 0.39\% and 0.41\% for pre-activation ResNet on SVHN, CIFAR-10 and CIFAR-100 datasets, respectively. This effect can be explained as follows: while the largest entropy is theoretically leading to the highest information capacity, it is also forcing the network to use the highest variety of weights, therefore preventing it from learning different modifications of the same filter targeted at different patterns, which can be crucial in many classification tasks.
	
	On the ImageNet classification problem, our PyTorch implementation of binary ResNet-18 and AlexNet models provided the same state-of-the-art accuracy (Table~\ref{table:bin:accuracy}) as the DoReFa-Nets~\cite{Zhou}
	with 4-bit activations: 59.2\% and 53.0\%, respectively. Our vanilla binary versions with full-precision external layers outperform 4-bit DenseNet-121 (4.1MB) and 6-bit Inception~v3 (20.4MB) with 8-bit activations that are showing the accuracy of 63.0\% and 72.5\%, respectively~\cite{Zhao}. Compared to the DenseNet-45 (7.4MB) with binary weights/activations and an accuracy of 63.7\%~\cite{Bethge}, we provide a more precise while 5.7$\times$ smaller DenseNet-121.	
	
	By adding the information loss penalty with $H_e=0.97$ and $\lambda = 10^{-4}$, the top-1 classification accuracy was drastically improved to 67.14\%, 61.36\%, 73.83\% and 54.07\% with DenseNet, ResNet, Inception and AlexNet models, respectively. While this performance still leaves room for improvement when compared to	full-precision models, it shows a significant advantage over the previously proposed attempts to improve the accuracy of the binarized networks with a fixed architecture and 4-bit activations on the ImageNet dataset.
	
	Figure~\ref{fig:curves} demonstrates that the information loss penalty	does not cause the training curve to be significantly altered, though it consistently shows a higher level of accuracy during the entire training. The reason for this is the stabilization of the information entropy in the binary network, reflected by a similar distribution of the real-valued and binary weights (Fig.~\ref{fig:weighs}) in the network's filters. As one can see, a higher level of $H$ corresponds not only to a lower skewness, but also to smaller absolute values of weights, which shows a similarity between the information loss penalty and the classical regularization methods such as $L_1$ and $L_2$ norms.
			
	\section{Conclusion}
	\label{Conclusion}
	In this paper, we report on the first attempt to control information capacity of deep binary networks with a new regularization technique~--- information loss penalty. By maintaining an appropriate level of information entropy in every convolutional filter and providing a stable information capacity of the entire binary network, the proposed algorithm provides better generalization ability and improves the accuracy in classification tasks. The main parameter of the algorithm~--- expected information entropy,~--- provides an opportunity to determine the optimal information capacity level and force the network to maintain this level throughout the entire learning process.
	By controlling information entropy of the binary network, we outperformed the existing state-of-the-art binary solutions using 4-bit activation functions and got closer to the prediction accuracy of their 32-bit counterparts on SVHN, CIFAR, and ImageNet machine learning datasets.
	
	\section*{Acknowledgement}\sm{We would like to thank Dr. Alexander Nikolaevich Filippov form Russian Research Center of Huawei Technologies for insightful discussions.}

\end{document}